%% file: eccv2022arxiv.tex
\documentclass[runningheads]{llncs}
\usepackage{graphicx}

\usepackage{tikz}
\usepackage{comment}
\usepackage{amsmath,amssymb,amsfonts} 
\usepackage{color}
\usepackage{xcolor}

\usepackage[accsupp]{axessibility}  

\begin{document}
\pagestyle{headings}
\mainmatter

\title{\raisebox{-0.16cm}{\includegraphics[scale=0.14]{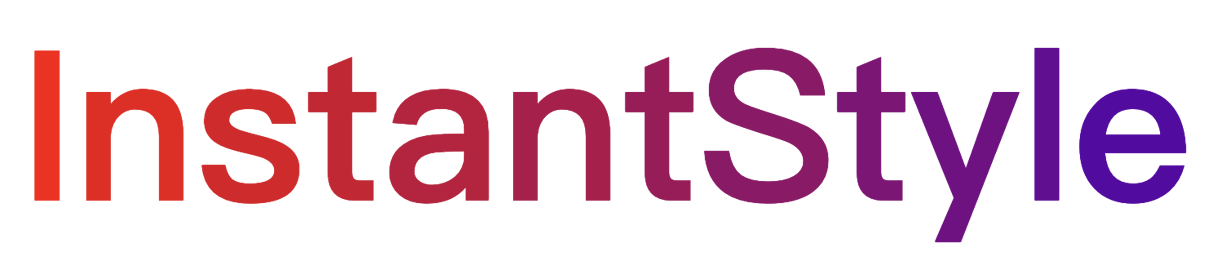}}: Free Lunch towards Style-Preserving in Text-to-Image Generation} 

\titlerunning{InstantStyle: Free Lunch towards Style-Preserving Generation}

\author{Haofan Wang \and
Matteo Spinelli \and
Qixun Wang \and
Xu Bai \and \\
Zekui Qin \and
Anthony Chen
}

\authorrunning{Wang et al.}
%


\institute{
InstantX Team \\
\email{\{haofanwang.ai@gmail.com\}\\
\textcolor{magenta}{\url{https://instantstyle.github.io}}
}
}
\maketitle

\begin{figure}[!h]
  \centering
  \includegraphics[width=0.85\textwidth]{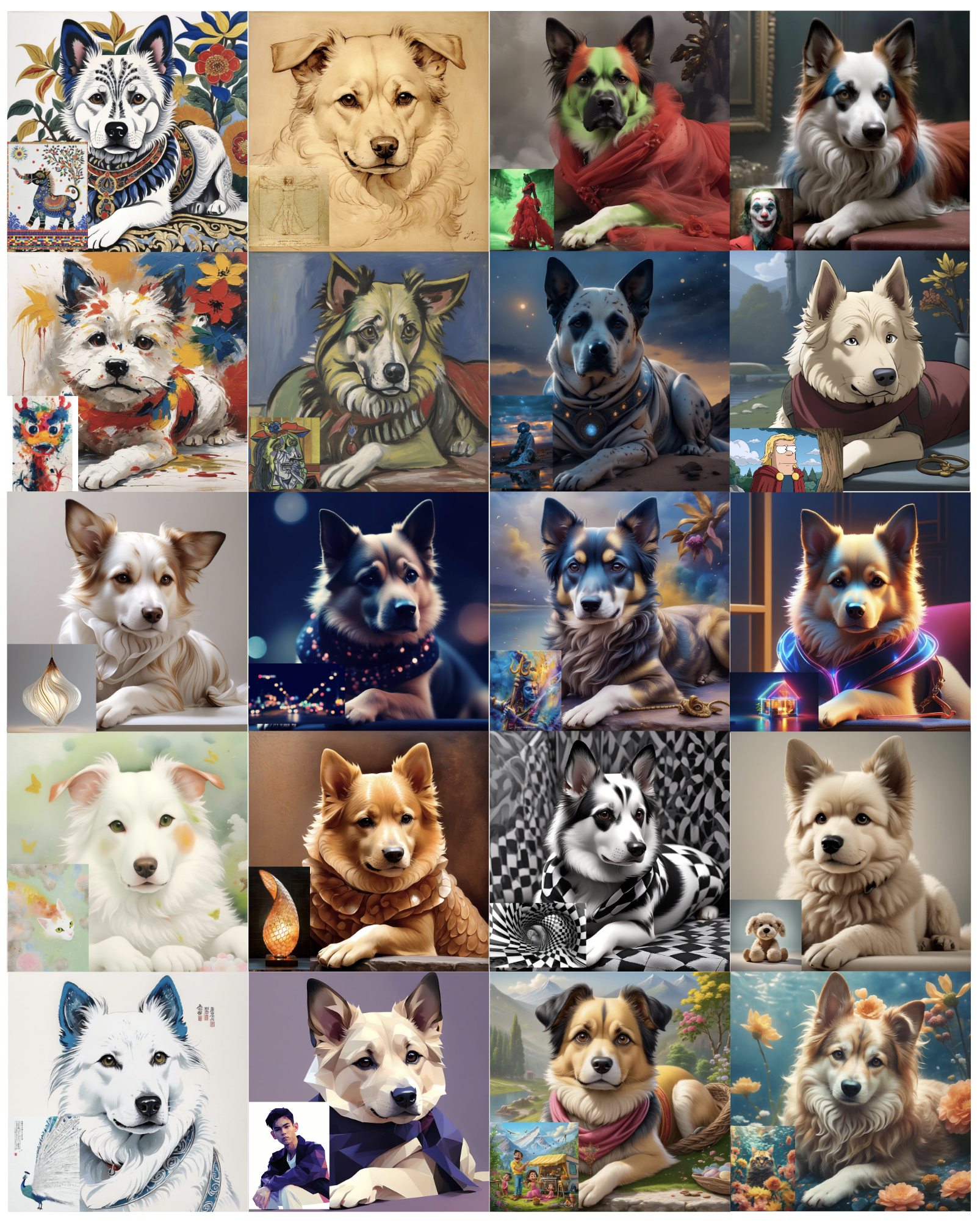}
  \caption{Stylized samples from InstantStyle.}
  \label{fig:0}
\end{figure}

\newpage

\begin{abstract}
Tuning-free diffusion-based models have demonstrated significant potential in the realm of image personalization and customization. However, despite this notable progress, current models continue to grapple with several complex challenges in producing style-consistent image generation. Firstly, the concept of style is inherently underdetermined, encompassing a multitude of elements such as color, material, atmosphere, design, and structure, among others. Secondly, inversion-based methods are prone to style degradation, often resulting in the loss of fine-grained details. Lastly, adapter-based approaches frequently require meticulous weight tuning for each reference image to achieve a balance between style intensity and text controllability. In this paper, we commence by examining several compelling yet frequently overlooked observations. We then proceed to introduce InstantStyle, a framework designed to address these issues through the implementation of two key strategies: 1) A straightforward mechanism that decouples style and content from reference images within the feature space, predicated on the assumption that features within the same space can be either added to or subtracted from one another. 2) The injection of reference image features exclusively into style-specific blocks, thereby preventing style leaks and eschewing the need for cumbersome weight tuning, which often characterizes more parameter-heavy designs.Our work demonstrates superior visual stylization outcomes, striking an optimal balance between the intensity of style and the controllability of textual elements. Our codes will be available at \textcolor{magenta}{https://github.com/InstantStyle/InstantStyle}.

\end{abstract}

\input{Sections/Introduction}
\input{Sections/Related}
\input{Sections/Method}

\input{Sections/Experiments}
\input{Sections/Conclusion}

\clearpage
%
%
\bibliographystyle{splncs04}
\bibliography{egbib}



\end{document}

%% file: Sections/Introduction.tex
\section{Introduction}


\begin{figure}[h]
  \centering
  \includegraphics[width=\textwidth]{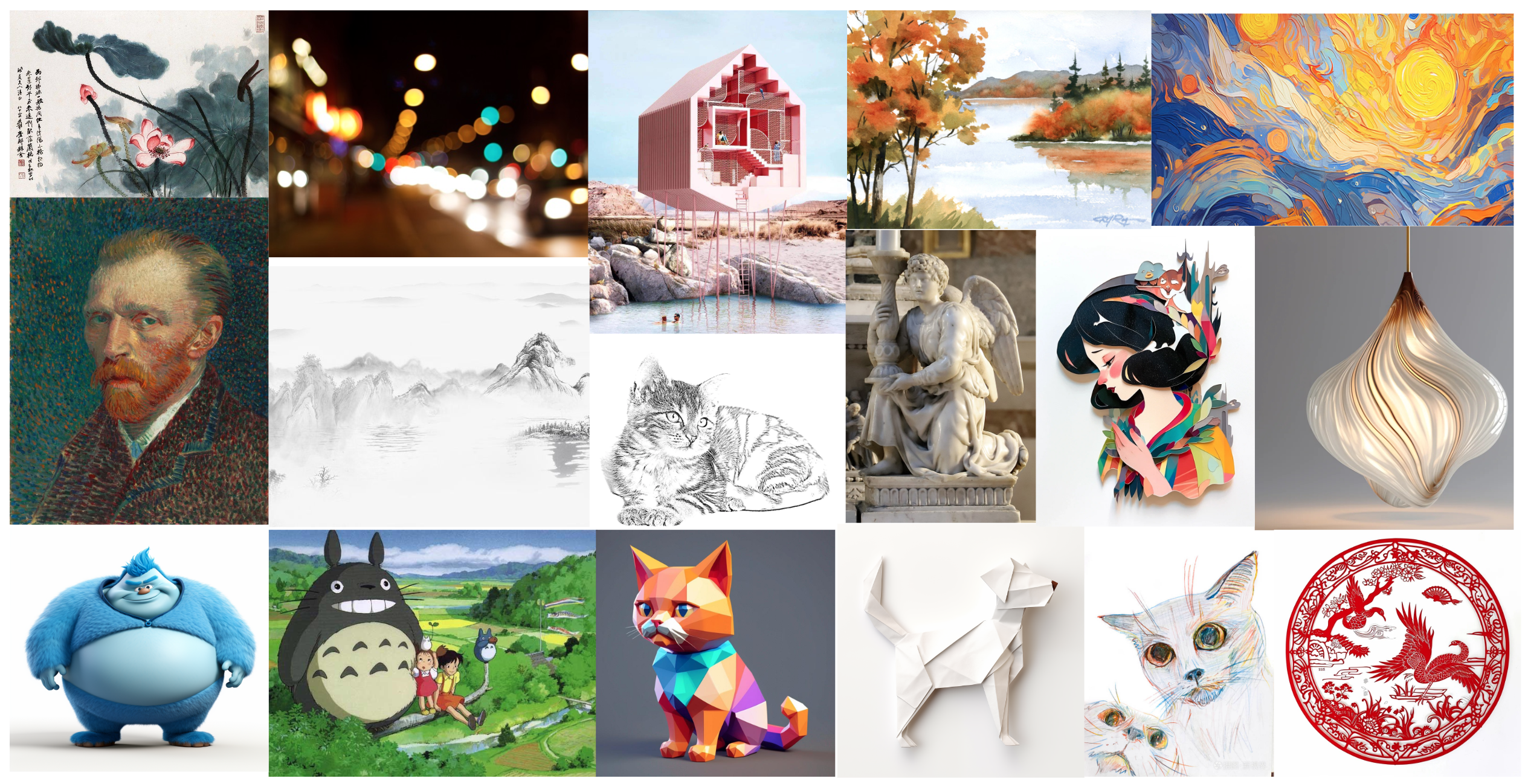}
  \caption{\textbf{Diversity of image styles.} A curated gallery of stylized images reveals the vast diversity and complexity of styles present in the real world, which are often challenging to define and categorize. While certain artistic styles, like ink and watercolor, have well-established criteria, the majority of natural images consist of a multitude of elements. This complexity poses a significant challenge for the task of style transfer, as it requires the intricate replication and integration of these elements into a new image while maintaining the original content's essence.}
  \label{fig:1}
\end{figure}

Diffusion-based text-to-image generative models have achieved notable advancements, in the areas of personalization and customization, particularly in consistent generation tasks such as identity preservation\cite{ye2023ip,wang2024instantid}, object customization\cite{chen2023anydoor}, and style transfer\cite{hertz2023style,sohn2023styledrop,jeong2024visual}. Despite these successes, style transfer remains a challenging task due to the underdetermined nature of style, which includes a variety of elements like color, material, atmosphere, design, and structure. The goal of style transfer, or stylized image generation, is to apply the specific style from a given reference image or subset to a target content image. The multifaceted attribute of style makes it difficult to collect stylized datasets, represent style accurately, and evaluate the success of the transfer.

Previous works\cite{hu2021lora,gal2022textual,ruiz2023dreambooth} have often involved fine-tuning diffusion models on a dataset of images that share a common style, which is both time-consuming and limited in its generalizability to real-world scenarios where it is difficult to gather a subset of images with a unified style. Recently, there has been a surge in interest in developing tuning-free approaches\cite{ye2023ip,hertz2023style,jeong2024visual,qi2024deadiff} for stylized image generation. These innovative methods can be broadly categorized into two distinct groups: 1) Adapter-free\cite{hertz2023style,jeong2024visual}: This category of methods leverages the power of self-attention within the diffusion process. By utilizing a shared attention operation, these techniques extract essential features such as keys and values directly from a given reference style image. This allows for a more streamlined and focused approach to image generation, as it draws directly from the stylistic elements present in the reference. 2) Adapter-based\cite{ye2023ip}: In contrast, adapter-based methods incorporate a lightweight model designed to extract detailed image representations from the reference style image. These representations are then skillfully integrated into the diffusion process via cross-attention mechanisms. This integration serves to guide the generation process, ensuring that the resulting images are aligned with the desired stylistic nuances of the reference.

Despite their promise, these tuning-free methods face several challenges. The Adapter-free approach involves an exchange of the Key and Value within the self-attention layer and pre-caches the K and V matrices derived from the reference style image. When applied to natural images, this method requires the inversion of the image back to latent noise through techniques like DDIM inversion\cite{song2020denoising} or similar methods. However, this inversion process can result in the loss of fine-grained details such as texture and color, thereby diminishing the style information in the generated images. Moreover, this additional step is time-consuming, which can be a significant drawback in practical applications. As for the Adapter-based method, the primary challenge lies in striking the right balance between style intensity and content leakage. Content leakage occurs when an increase in style intensity leads to the appearance of non-style elements from the reference image in the generated output. Essentially, the difficulty is in effectively separating style from content within the reference images. Some approaches aim to address this by constructing paired datasets where the same object is represented in multiple styles, facilitating the extraction of disentangled style and content representations. However, due to the inherently underdetermined nature of style, the creation of large-scale paired datasets is both resource-intensive and limited in the diversity of styles it can capture. This limitation, in turn, restricts the generalization capabilities of these methods, as they may not be as effective when applied to styles outside of the dataset.

In view of these limitations, we introduce a novel tuning-free mechanism (InstantStyle) based on existing adapter-based method, which can also be seamlessly integrated into the other existing attention-based injection method and effectively achieve the decoupling of style and content. More specifically, we introduce two simple but effective ways to complete the decoupling of style and content, thereby achieving better style migration, without need to build paired datasets or introduce additional modules to achieve decoupling. (1) Although previous adapter-based methods have widely used CLIP\cite{clip} as an image feature extractor, few works have considered feature decoupling within the feature space. Compared with the underdetermination of style, content is usually easier to describe with text. Since text and images share a feature space in CLIP, we found that a simple subtraction operation of image features and content text features can effectively reduce content leakage; (2) Inspired by previous works\cite{voynov2023p+,jeong2024visual,frenkel2024implicit}, we found that in the diffusion model, there is a specific layer responsible for the injection of style information. By injecting image features only into specific style blocks, the decoupling of style and content can be accomplished implicitly. With just these two simple strategies, we solved most of the content leakage problems while maintaining the strength of the style.

In summary, we share several valuable insights and present InstantStyle that employs two straightforward yet potent techniques for achieving an effective disentanglement of style and content from reference images. It is tuning-free, model independent, and pluggable with other attention-based feature injection works, showing excellent style transfer performance and igniting huge potential for downstream tasks and other domains.





%% file: Sections/Related.tex
\section{Related Work}

\subsection{Text-to-image Diffusion Models} 
Text-to-image diffusion models\cite{rombach2022high,zhang2023adding,li2024playground} have emerged as a pivotal approach in the realm of generative visual content, showcasing an unparalleled ability to create high-quality images that are aligned with textual descriptions. These models are distinguished by their use of a diffusion process that is intricately conditioned on the provided text via cross-attention layers, ensuring that the generated images are not only visually coherent but also semantically consistent with the input descriptions. Among these works, stable diffusion\cite{rombach2022high,podell2023sdxl} is the most typical representative, in which the diffusion process is completed in the latent space. Although the DiT\cite{peebles2023scalable} architecture has recently begun to gradually replace the UNet\cite{ronneberger2015u} architecture, there is no doubt that its emergence has promoted the development and prosperity of the open source ecosystem and greatly promoted the explosion of downstream applications.

\subsection{Stylized Image Generation}
Stylized image generation, also often called style transfer, its goal is to transfer the style of a reference image to the target content image. Previous customization works\cite{hu2021lora,gal2022image,ruiz2023dreambooth} usually fine-tine the diffusion model on a set of images with the same style, which is time-costing and cannot well generalize to real world domain where a subset with shared style is hard to collect. Recently, there has been
a surge in interest in developing tuning-free approaches\cite{sohn2023styledrop,ye2023ip,wang2023styleadapter,hertz2023style,jeong2024visual,qi2024deadiff,frenkel2024implicit} for stylized image generation. These works use lightweight adapters to extract image features and inject them into the diffusion process through self-attention or cross-attention. IP-Adapter\cite{ye2023ip} and Style-Adapter\cite{wang2023styleadapter} share the same idea where a decoupled cross-attention mechanism is introduced to separate cross-attention layers for text features and image features. However, they suffer from content leakage more or less as suggested in \cite{jeong2024visual}. StyleAlign\cite{hertz2023style} and Swapping Self-Attention\cite{jeong2024visual} swap the key and value features of self-attention
block in an original denoising process with the ones from a reference denoising process. But for real world images, they requires an inversion to turn image back to a latent noise, leading to a loss of fine-grained details such as texture and color, thereby diminishing the style information in
the generated images. DEADiff\cite{qi2024deadiff} aims to extract disentangled representations of content and style utilizing a paired dataset and Q-Former\cite{li2023blip}. However, due to the inherently underdetermined nature of style,
the construction of large-scale paired datasets is resource-intensive and limited in the diversity of styles. For style transfer, we support lightweight modules such as IP-Adapter\cite{ye2023ip} because of its portability and efficiency. The only problem is how to complete the decoupling of content and style in images.

\subsection{Attention Control in Diffusion Models}
As shown in \cite{hertz2022prompt}, self-attention and cross-attention blocks within diffusion process determine different attributes, such as spatial layout and content of the generated images. Image editing approaches \cite{brooks2023instructpix2pix,mokady2023null,tumanyan2023plug} apply attention control to enable structure-preserving image edit. $P+$\cite{voynov2023p+} demonstrates that different cross-attention layers in the diffusion U-Net express distinct responses to style and semantics, provides greater disentangling and control over image synthesis. Swapping Self-Attention \cite{jeong2024visual} reveals that upblocks in UNet appropriately reflect the style elements, while  bottleneck and downblocks cause content leakage. DEADiff\cite{qi2024deadiff} introduces a disentangled conditioning mechanism that conditions the coarse
layers with lower spatial resolution on semantics, while the fine layers with higher spatial resolution are conditioned on the style. A more recent work, B-LoRA\cite{frenkel2024implicit} find that
jointly learning the LoRA weights of two specific blocks implicitly separates the style and content components of a single image. Our work is mostly inspired by these works, and we aim to identify the most style-relevant layers for disentangling content and style in style transfer.

%% file: Sections/Method.tex
\section{Methods}
    
\subsection{Motivations}
\label{sec:preliminary}

\subsubsection{The definition of style is under determined.} Previous consistency tasks, such as ID consistency generation\cite{wang2024instantid}, can measure the quality of generation through the similarity of faces. There are objective evaluation metrics. Although these quantitative indicators do not fully represent fidelity, they can at least be measured through user study. However, for style consistency, such metrics are lacking. The core here is that the style of the picture lacks a unified definition. In different scenes, the meaning of style is significantly different. For example, style can be the ink elements in a Chinese landscape paintings, or it can be the difference in pigments such as watercolor and oil painting, or it can be layout, material, atmosphere, etc. Figure \ref{fig:1} shows a style gallery, covering film, painting, design, architecture, etc. More extremely, a round red hollow window grille as shown in the right bottom of Figure \ref{fig:1}, which contains shapes, colors, and designs, making it difficult to distinguish the style. In shorts, style is usually not a single element, but a combination of multiple complex elements. This makes the definition of style a relatively subjective matter and not unique, with multiple reasonable interpretations of the same style.

The consequence of this is that it is difficult or even impossible to collect data on the same style of pairings on a large scale. There have been some previous works\cite{qi2024deadiff} that used large language models to generate style descriptions, and then used closed-source text-to-image generation models, such as Midjourney\footnote{https://www.midjourney.com/}, to generate images of specific styles. However, there are several problems with this. First, there is a lot of noise in the original generated data, and since the subdivision styles are difficult to distinguish, cleaning will be quite difficult. Second, the number of styles is often limited because many of the more fine-grained styles cannot be clearly described through text, and the types of styles are limited by Midjourney's capabilities, which can lead to limited diversity.

\begin{figure}[h]
  \centering
  \includegraphics[width=\textwidth]{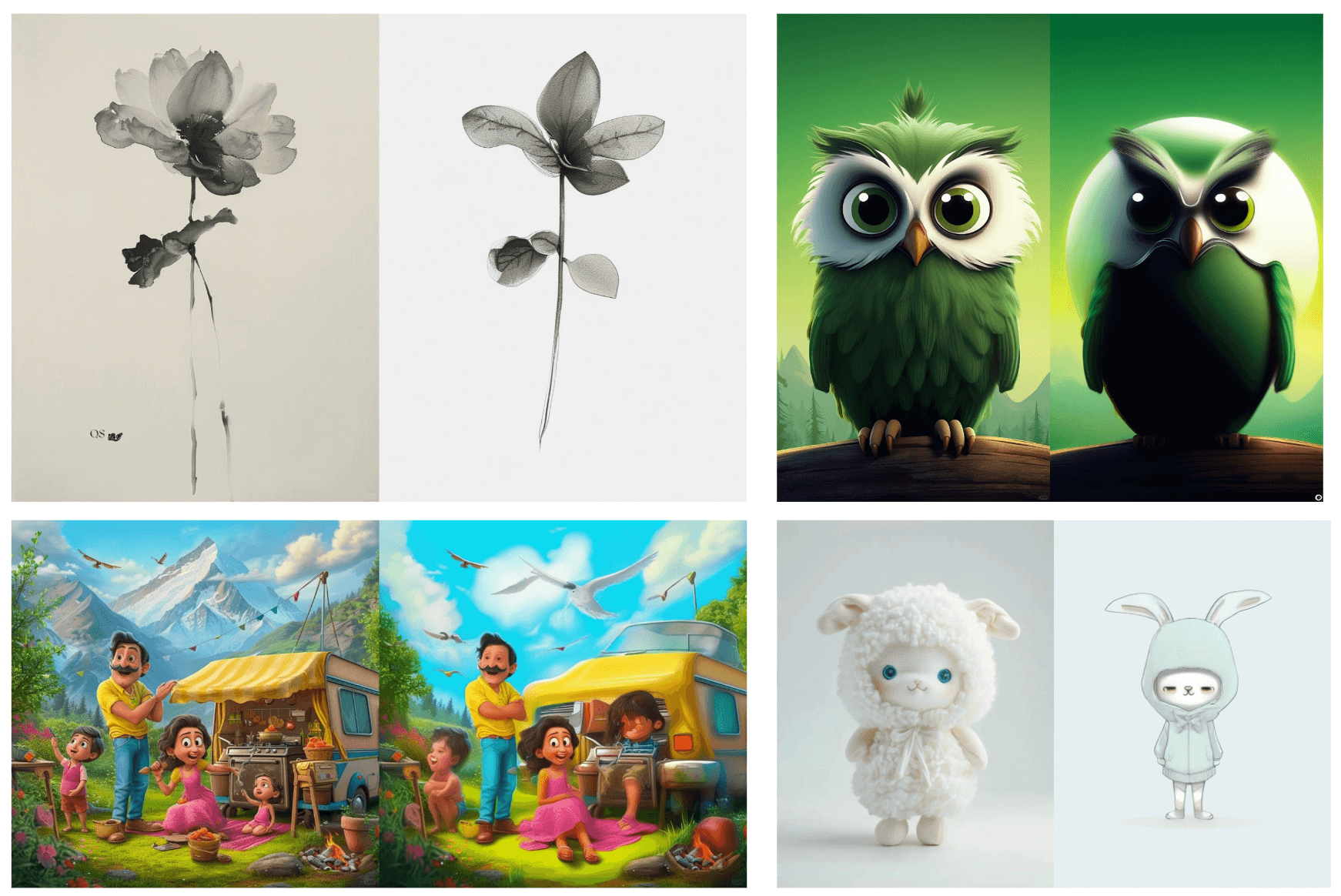}
  \caption{\textbf{DDIM inversion with style degradation.} For real world image (left), DDIM inversion reconstruction (right) is inadequate to retain fine-grained details, which can be necessary for styles.}
  \label{fig:2}
\end{figure}

\begin{figure}[h]
  \centering
  \includegraphics[width=\textwidth]{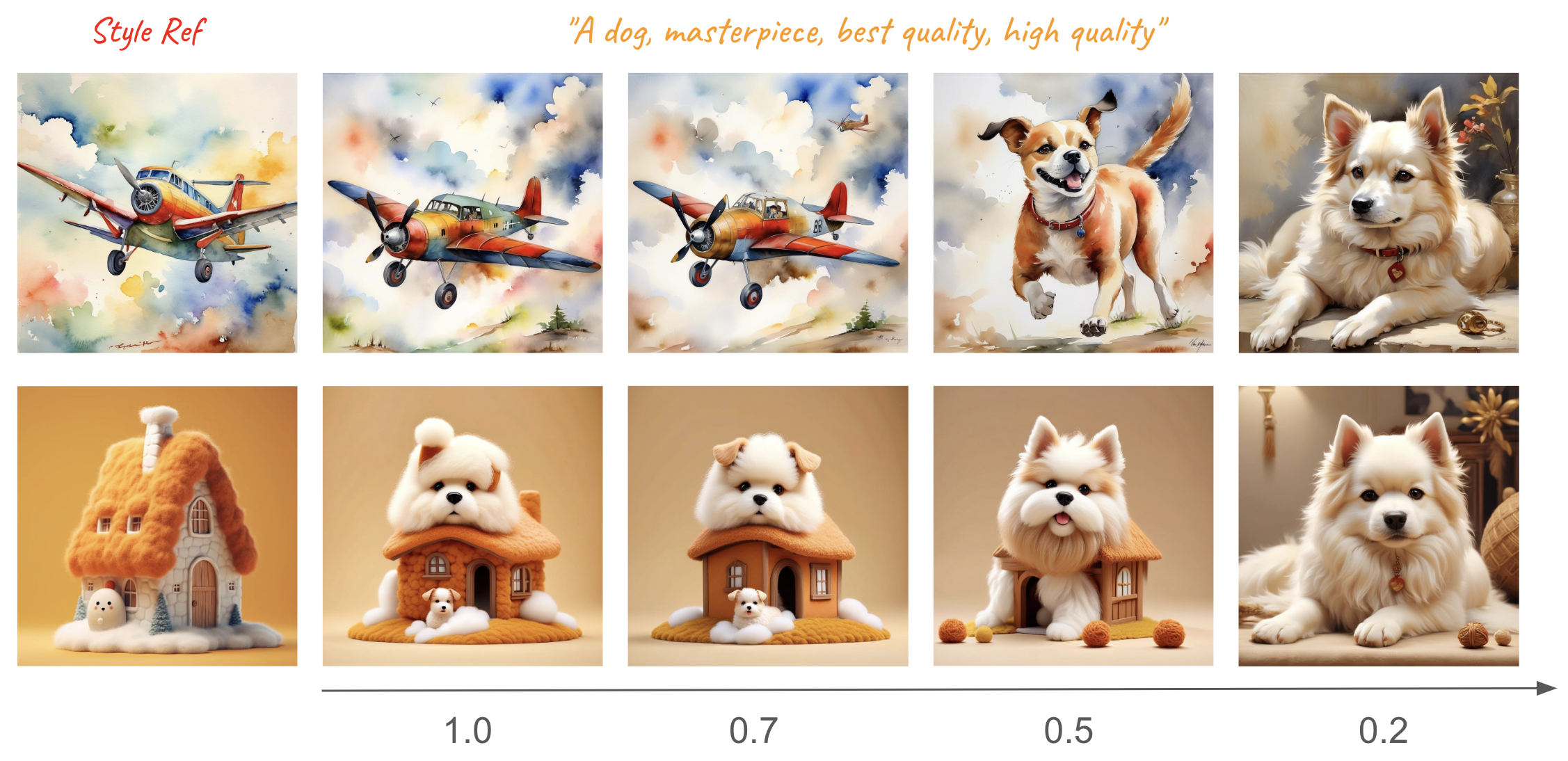}
  \caption{\textbf{Trade-off between style strength and content leakage.} Given a style reference image, the strength of style image affects the generated result. For high strength, the text controllability is
   damaged and it often comes with content leakage. For low strength, the style information cannot well guide the generation.}
  \label{fig:3}
\end{figure}

\begin{figure}[!h]
  \centering
  \includegraphics[width=\textwidth]{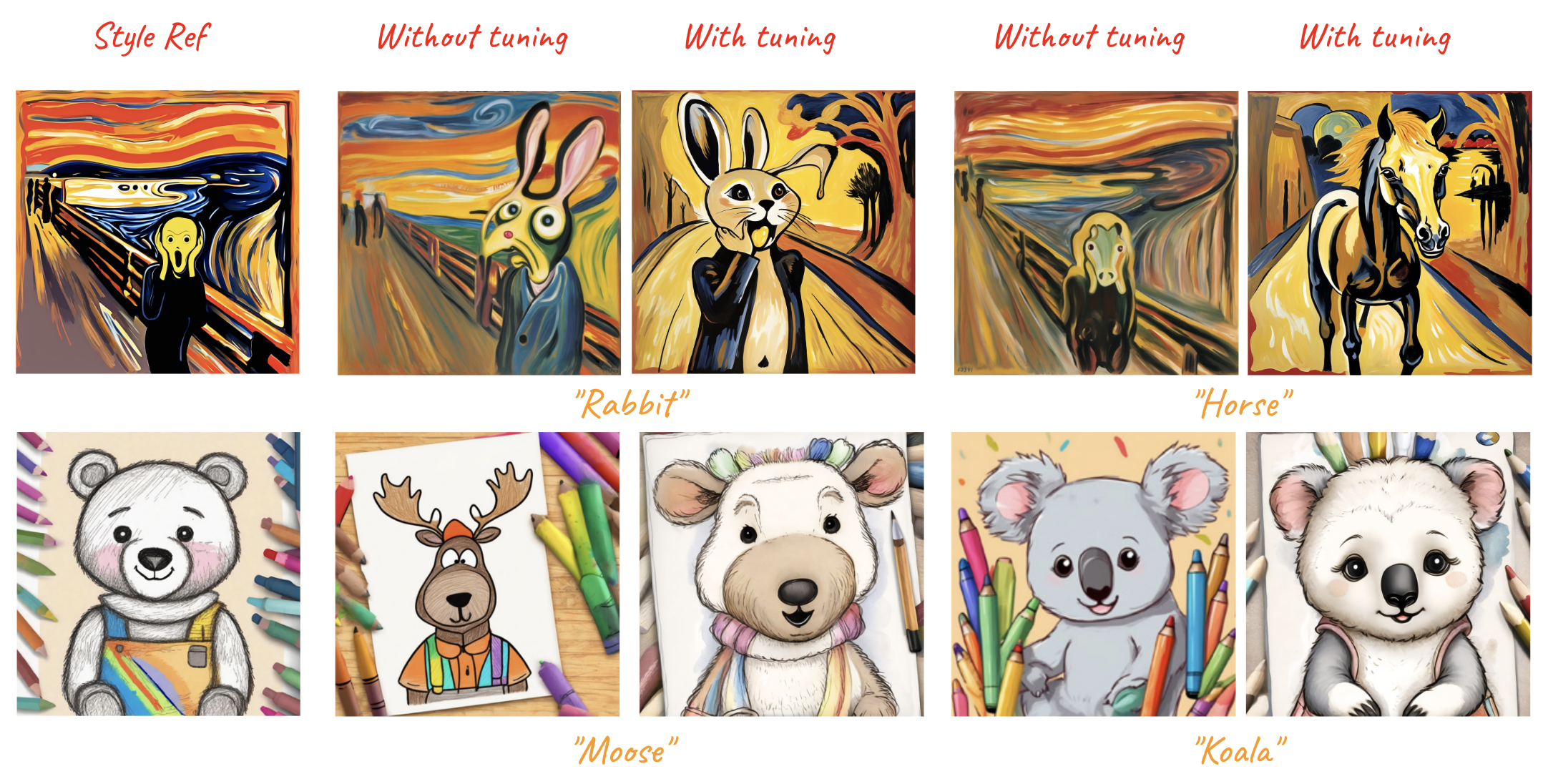}
  \caption{\textbf{Opportunistic visual comparison.} Given a style reference image, weight tuning plays an important role on style strength and content leakage.}
  \label{fig:4}
\end{figure}

\begin{figure}[h]
  \centering
  \includegraphics[width=\textwidth]{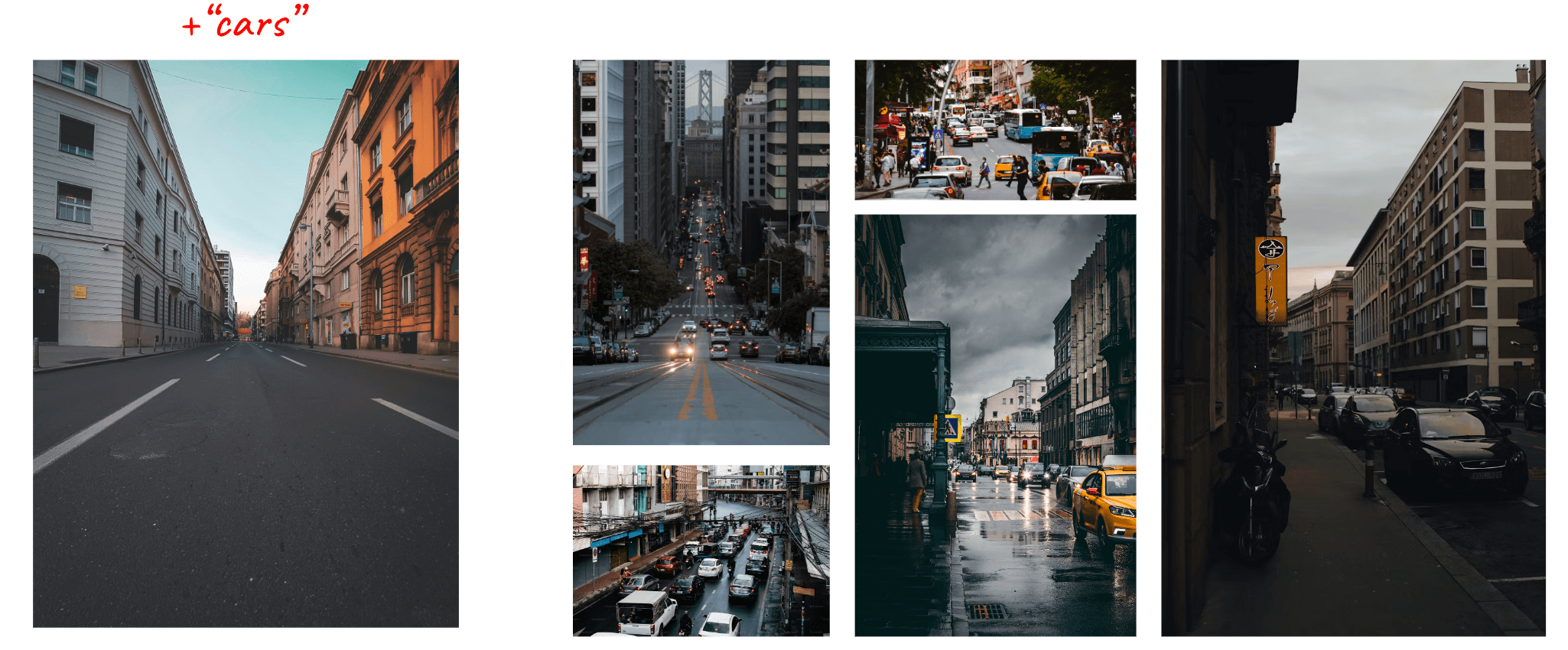}
  \caption{\textbf{Multi-modality image retrival using CLIP.} Given a query image of empty street and a query prompt `cars', CLIP supports joint query search from Unsplash.}
  \label{fig:5}
\end{figure}

\begin{figure}[h]
  \centering
  \includegraphics[width=\textwidth]{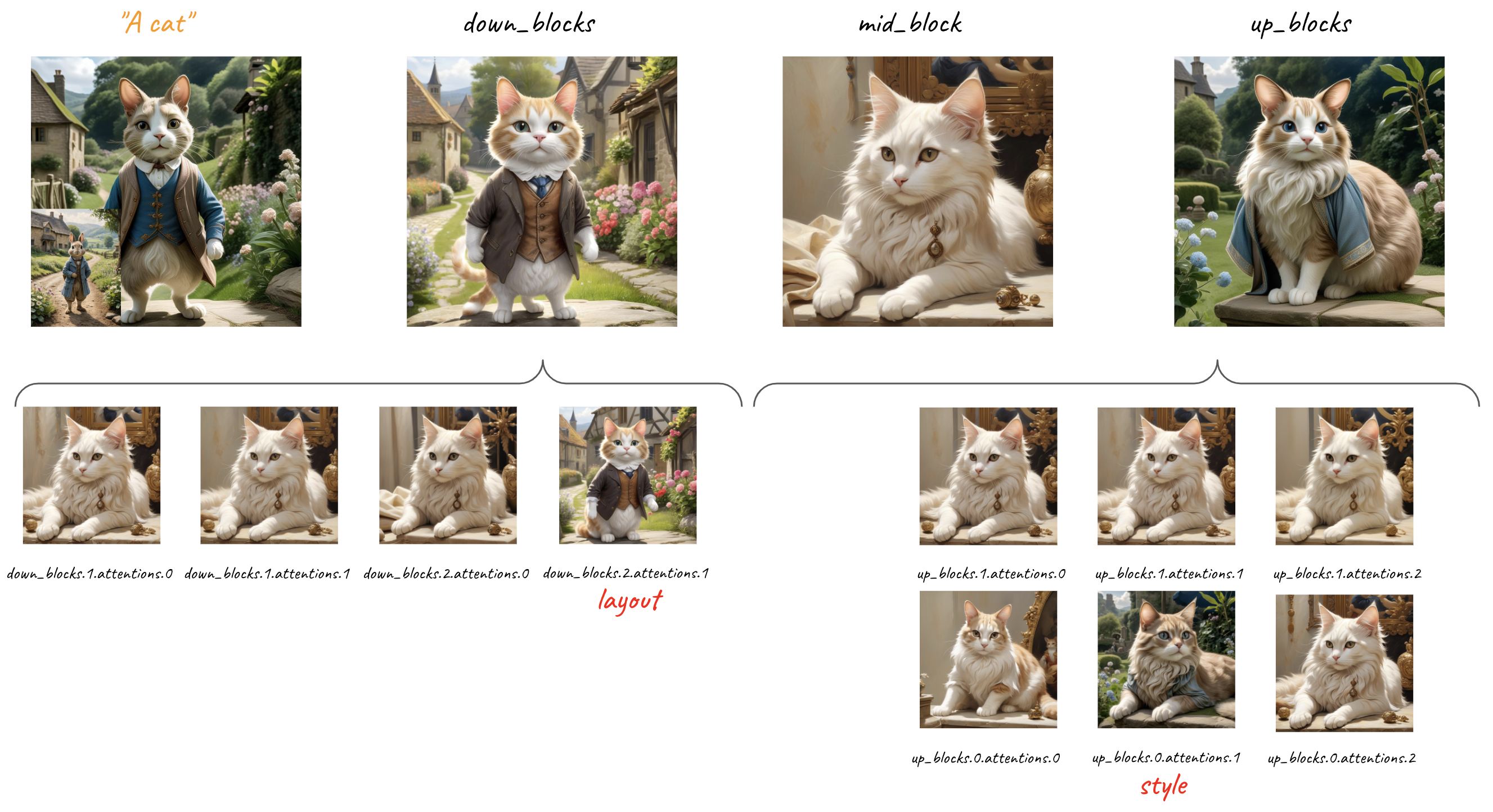}
  \caption{\textbf{Fine-grained Analysis over Attention Blocks.} We conduct comprehensive experiments on IP-Adapter (SDXL) to analyse the effect of each attention layer for image feature. We are surprised to find that up\_blocks.0.attentions.1 and down\_blocks.2.attentions.1 are the most representative layers, controlling style and spatial layout respectively. As for whether layout is a style, it varies from person to person.}
  \label{fig:6}
\end{figure}

\subsubsection{Style degradation caused by inversion.} 
In the inversion-based method, given an input reference image and text description, DDIM inversion technique is adopted over the image to get the inverted diffusion trajectory $x_{T}$, $x_{T-1}$ ... $x_{0}$, transforming an image into a latent noise representation by approximating the inversion equation. Then, starting from $x_{T}$ and a new set of prompts, these methods generate new content with an aligned style to the input. However, as shown in Figure \ref{fig:2}, DDIM inversion for real images is unstable as it relies on local linearization assumptions, which result in the propagation of errors, leading to incorrect image reconstruction and loss of content. Intuitively, the inverted result will lose a lot of fine-grained style information, such as textures, materials, etc. In this case, using pictures that have lost style as a guidance condition obviously cannot effectively achieve style transfer. In addition, the inversion process is also parameter sensitive and causes the generation speed to be very slow.

\subsubsection{Trade-off between style strength and content leakage.} As found in previous works\cite{jeong2024visual,qi2024deadiff}, there is also a balance in the injection of style conditions. If the intensity of the image condition is too high, the content may be leaked, while if the intensity is too low, the style may not be obvious enough. The core reason for this is that the content and style in the image are coupled, and due to the underdetermined attributes of style we mentioned above, the two are often difficult to decouple. Therefore, meticulous weight is usually required tuning for each reference image to balance style strength and text controllability.

\subsection{Observations}
\label{sec:preliminary}

\subsubsection{Adapter’s capabilities are underestimated.} Interestingly, we found that the evaluation of IP-Adapter's style transfer ability in most previous works was biased. Some of them\cite{jeong2024visual} use fixed parameters to claim that IP-Adapter is weakened in text control capabilities, while others emphasize the problem of content leakage. This can basically be attributed to the same reason, that is, the appropriate strength is not set. Since the IP-Adapter uses two sets of K and V to process text and images in cross-attention, and finally performs linear weighting, too high a strength will inevitably reduce the control ability of the text. At the same time, because the content and style in the image are not decoupled, this leads to the problem of content leakage. The simplest and most effective way to do this is to use a lower strength, and we have found that this solves the problems mentioned in most of the work, although we also agree that weight tuning is a very tricky thing and does not always work.

\subsubsection{Subtracted CLIP's embeddings as disentangled representation.} The original CLIP model aimed to unite image and text modalities within a shared embedding space using a contrastive loss, trained on large scale weakly-aligned text-image pairs. Most of previous adaptor-based methods use a pretrained CLIP image encoder model to extract image features from given image. Among these works, the global image embedding from the CLIP image encoder is commonly utilized as it can capture overall content and style of the image. Although we mentioned above that the problem of content leakage can be alleviated by reducing strength, for our task, this still raises content and style coupling issues in some cases. There have been some previous works by constructing paired style data sets to extract representations of style and content respectively, but they were limited by the diversity of styles. In fact, there are simpler and more effective ways motivated by image retrieval\footnote{https://github.com/haofanwang/natural-language-joint-query-search}. Don’t forget that CLIP’s feature space has good compatibility, and features in the same feature space can be added and subtracted, as shown in Figure \ref{fig:5}. The answer is actually obvious. Since pictures contain both content and style information, and compared to style, content can often be described by text, we can explicitly eliminate the content part of the picture features, assuming the new feature is still in CLIP space.

\subsubsection{The impact of different blocks is not equal.} In the era of convolutional neural networks, many studies have found that shallow convolutional layers will learn low-level representations, such as shape, color, etc., while high-level layers will focus on semantic information. In diffusion-based models, the same logic exists as found in \cite{frenkel2024implicit,voynov2023p+,qi2024deadiff}. Like text conditions, image conditions are generally injected through cross attention layer to guide generation. We found that different attention layers capture style information differently. In our experiments as shown in Figure \ref{fig:6}, we found that there are two special layers that play an important role in style preservation. To be more specific, we find up\_blocks.0.attentions.1 and down\_blocks.2.attentions.1 capture style (color, material, atmosphere) and spatial layout (structure, composition) respectively, the consideration of layout as a stylistic element is subjective and can vary among individuals.

\subsection{Methodology}
\label{sec:methodology}

\begin{figure}[h]
  \centering
  \includegraphics[width=\textwidth]{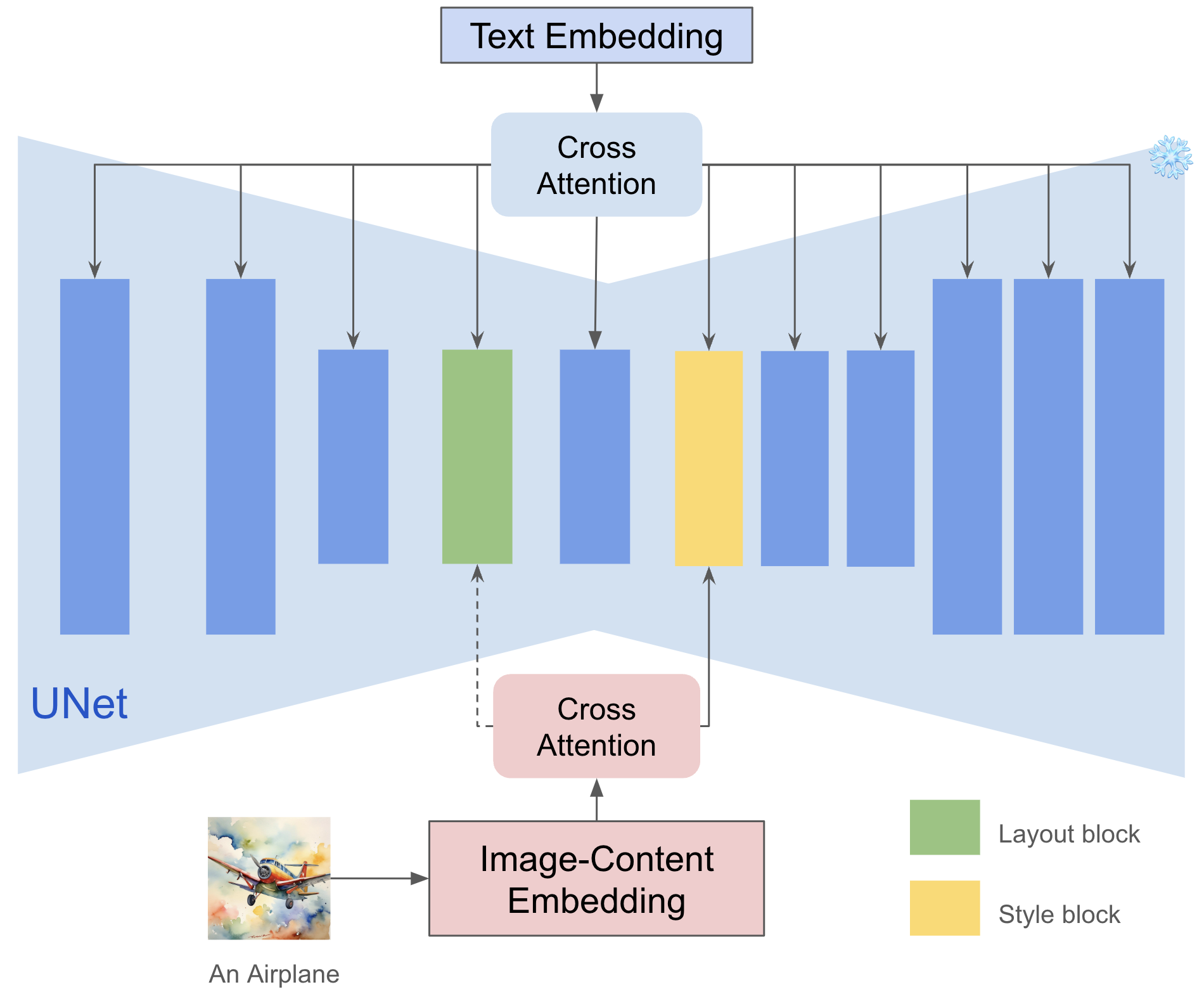}
  \caption{\textbf{Overview of IP-Adapter with InstantStyle.} There are 11 transformer blocks with SDXL, 4 for downsample blocks, 1 for middle block, 6 for upsample blocks. We find that the $4_{th}$ and the $6_{th}$ blocks are corresponding to layout and style respectively. Most time, the $6_{th}$ blocks is enough to capture style, the $4_{th}$ matters only when the layout is a part of style in some cases. In addition, you can also optionally use the characteristics of CLIP to explicitly subtract content from the feature space.}
  \label{fig:pipe}
\end{figure}

Based on our observations, in this section, we explain how to achieve decoupling of content and style through two simple and effective strategies. These two strategies do not conflict with each other and can be used separately for different models seamlessly as a free lunch. Figure \ref{fig:pipe} shows our pipeline.

\subsubsection{Separating Content from Image.} Instead of employing complex strategies to disentangle content and style from images, we take the simplest approach to achieving similar capabilities. Compared with the underdetermined attributes of style, content can usually be represented by natural text, thus we can use CLIP's text encoder to extract the characteristics of the content text as content representation. At the same time, we use CLIP's image encoder to extract the features of the reference image as we did in previous work. Benefit from the good characterization of CLIP global features, after subtracting the content text features from the image features, the style and content can be explicitly decoupled. Although simple, this strategy is quite effective in mitigating content leakage.

\subsubsection{Injecting into Style Blocks Only.} Empirically, each layer of a deep network captures different semantic information, as we mentioned in Figure \ref{fig:6}, the key observation in our work is that there exists two specific attention layers handling style. Specifically, we find up\_blocks.0.attentions.1 and down\_blocks.2.attentions.1 capture style (color, material, atmosphere) and spatial layout (structure, composition) respectively. More intuitively, the $4_{th}$ and the $6_{th}$ as shown in. Figure \ref{fig:pipe} for easy understanding. We can use them to implicitly extract style information, further preventing content leakage without losing the strength of the style. The idea is straightforward, as we have located style blocks, we can inject our image features into these blocks only to achieve style transfer seamlessly. Furthermore, since the number of parameters of the adapter is greatly reduced, the text control ability is also enhanced. This mechanism is applicable to other attention-based feature injection for editing or other tasks.

%% file: Sections/Experiments.tex
\section{Experiments}

\begin{figure}[!h]
  \centering
  \includegraphics[width=\textwidth]{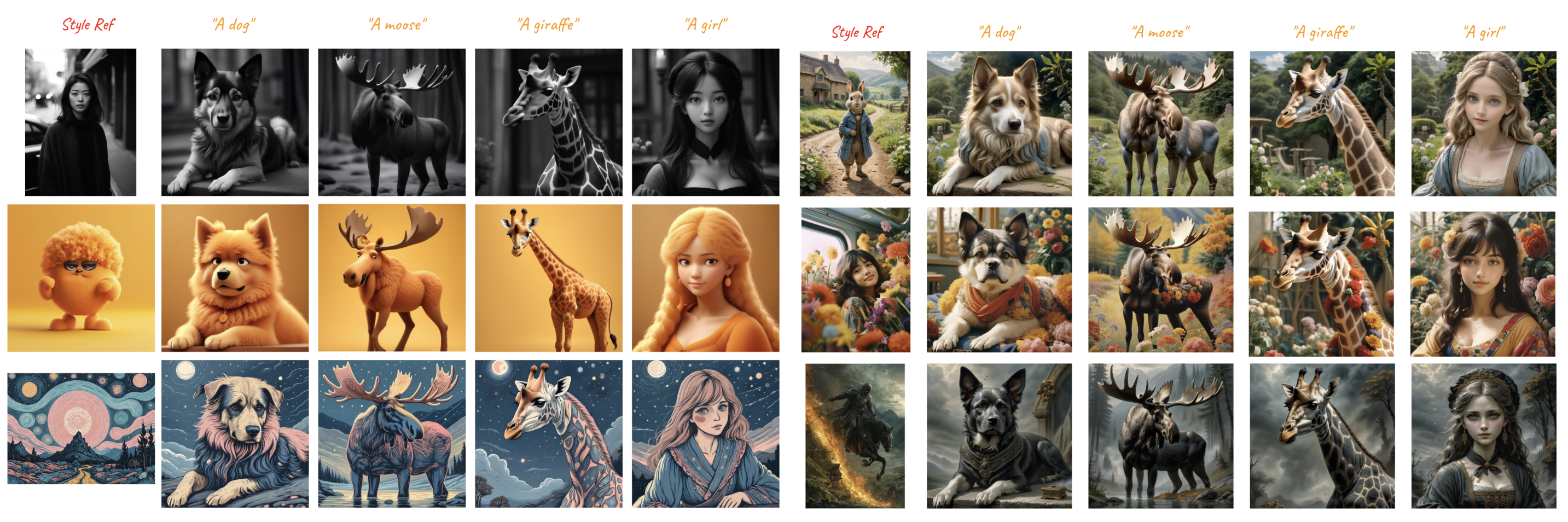}
  \caption{\textbf{Qualitative Results.} Given a single style reference image and different prompts, our work achieves high style-consistent generation.}
  \label{fig:7}
\end{figure}

\begin{figure}[!h]
  \centering
  \includegraphics[width=\textwidth]{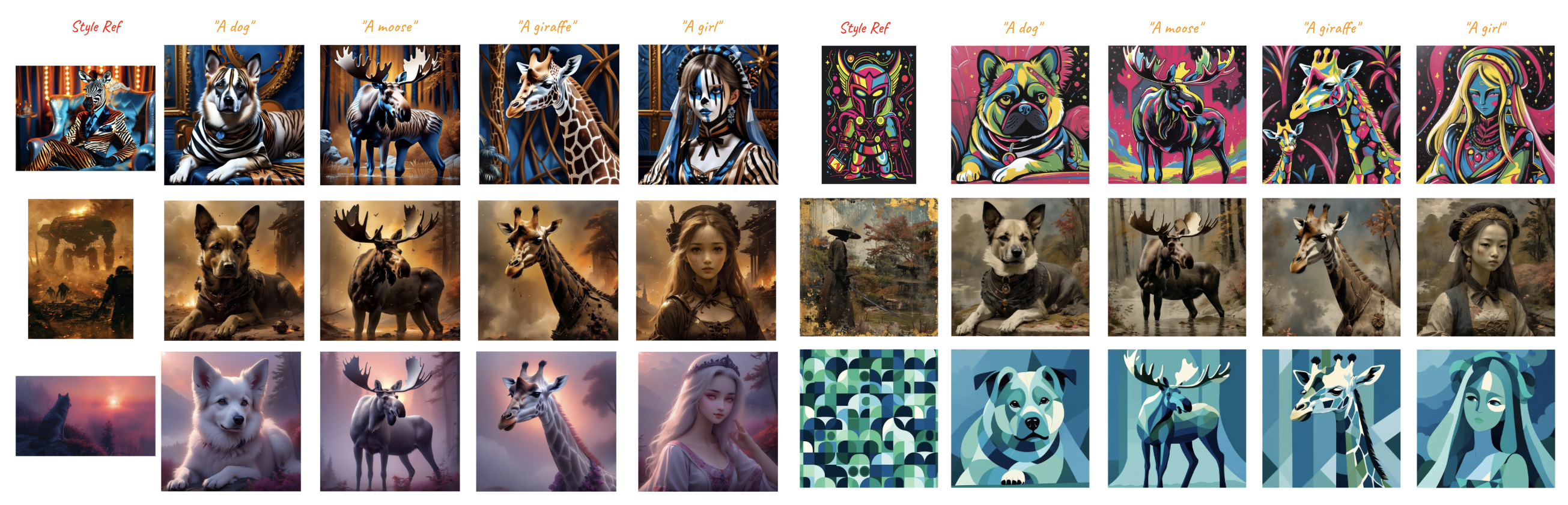}
  \caption{\textbf{More Qualitative Results.} Given a single style reference image and different prompts, our work achieves high style-consistent generation.}
  \label{fig:8}
\end{figure}

We have implemented our method on Stable Diffusion XL (SDXL). We use commonly adopted pretrained IP-Adapter as our exemplar to validate our methodology, but for image features, we mute out all blocks except for style blocks. We also trained IP-Adapter (SDXL 1.0) on 4M large scale text-image paired datasets from scratch following its official training setting, instead of training all blocks, in our cases, only style blocks will be updated. Interestingly, we find that these two settings achieve quite similar stylized results, thus for convenience, our following experiments are all based on pretrained IP-Adapter without further fine-tuning.

\subsection{Qualitative Results}

\subsubsection{Text-based image stylization.}

To verify the robustness and generalization capacities of InstantStyle, we conducted numerous style transfer experiments with various styles across different content.
Figure \ref{fig:7} and Figure \ref{fig:8} present the style transfer results. Since image information is only injected into style blocks, content leakage is greatly mitigated and careful weight tuning is not required. These results are not cherrypicked.

\subsubsection{Image-based image stylization.} 
\begin{figure}[!h]
  \centering
  \includegraphics[width=\textwidth]{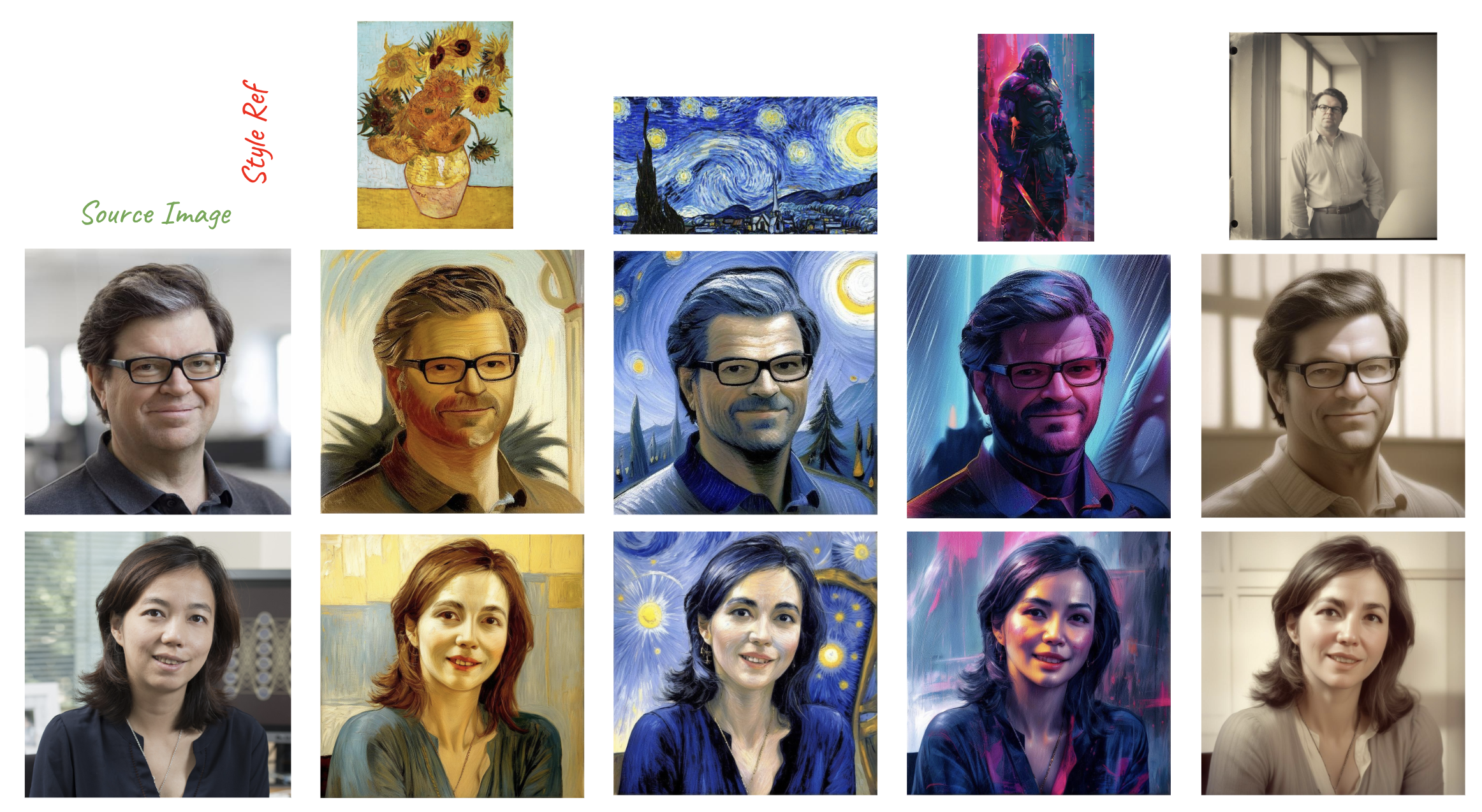}
  \caption{\textbf{Image-based image stylization results.} Given a single style reference image and different prompts, our work is compatible with ControlNets to achieve image stylization.}
  \label{fig:11}
\end{figure}

We also adopt ControlNet (Canny) to achieve image-based stylization with spatial control, the results are shown in Figure \ref{fig:11}.

\subsection{Comparison to Previous Methods}

For baselines, we compare our method to recent state-of-the-art stylization methods, including StyleAlign\cite{hertz2023style}, Swapping Self-Attention\cite{jeong2024visual}, B-LoRA\cite{frenkel2024implicit} and original IP-Adapter\cite{ye2023ip} with weight tuning. For B-LoRA, we train on single reference style image using the official training setting.

\begin{figure}[!h]
  \centering
  \includegraphics[width=\textwidth]{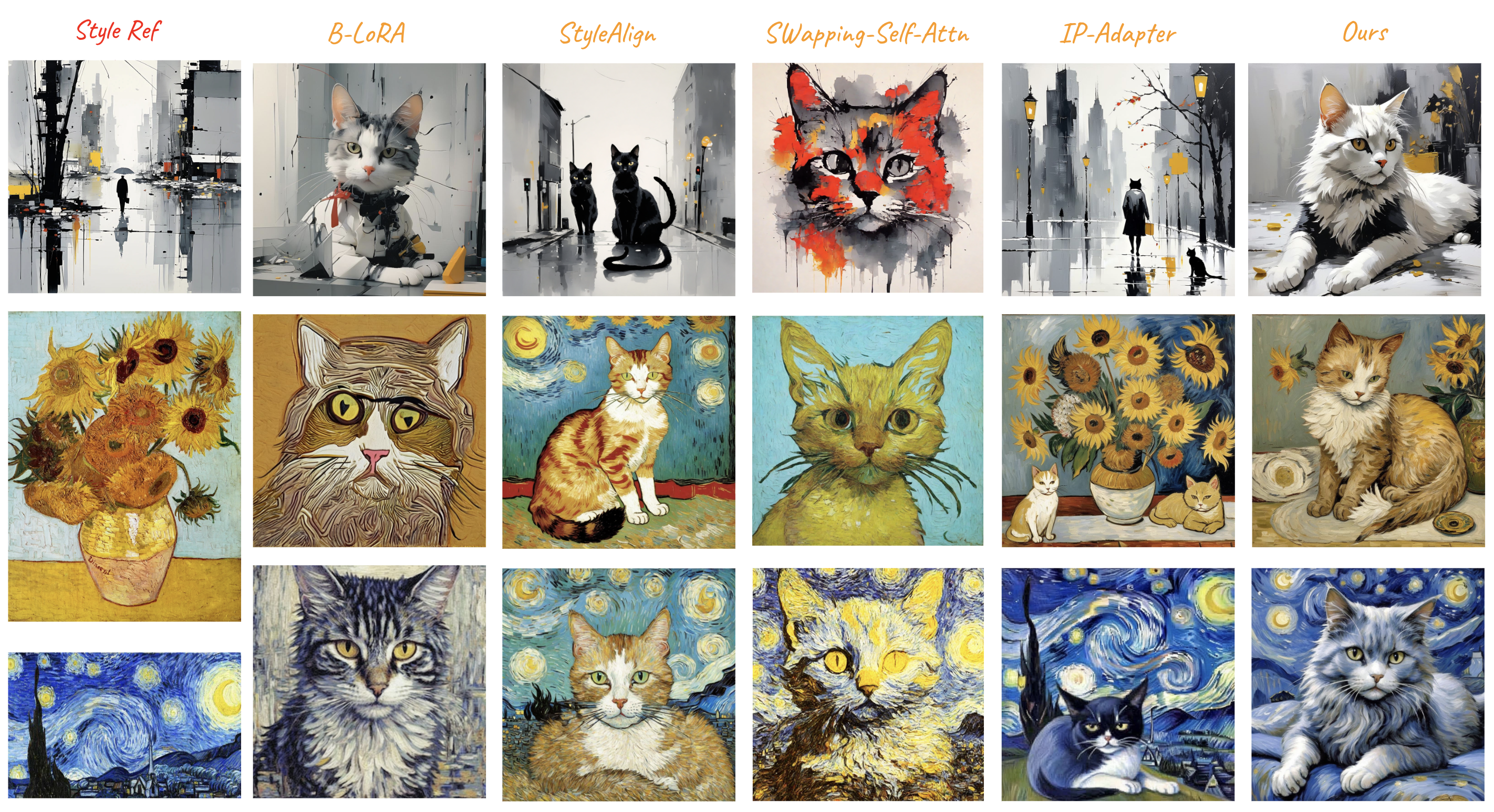}
  \caption{\textbf{Comparison to Previous Methods.} For previous works, we use their official implementations with manual weight tuning for fair comparison, as some of them are usually strength sensitive.}
  \label{fig:10}
\end{figure}

Frankly speaking, the style definition of each method is different from different perspectives, but in general, our method achieves the best visual effect.

\subsection{Ablation Study}
In this section, we study the effect of our proposed strategies respectively. We carefully tune the strength for the original IP-Adapter,  IP-Adapter with the content embedding subtracted. After applying style block injection, no tuning is adopted anymore, we fix the strength to 1.0.

\begin{figure}[!h]
  \centering
  \includegraphics[width=\textwidth]{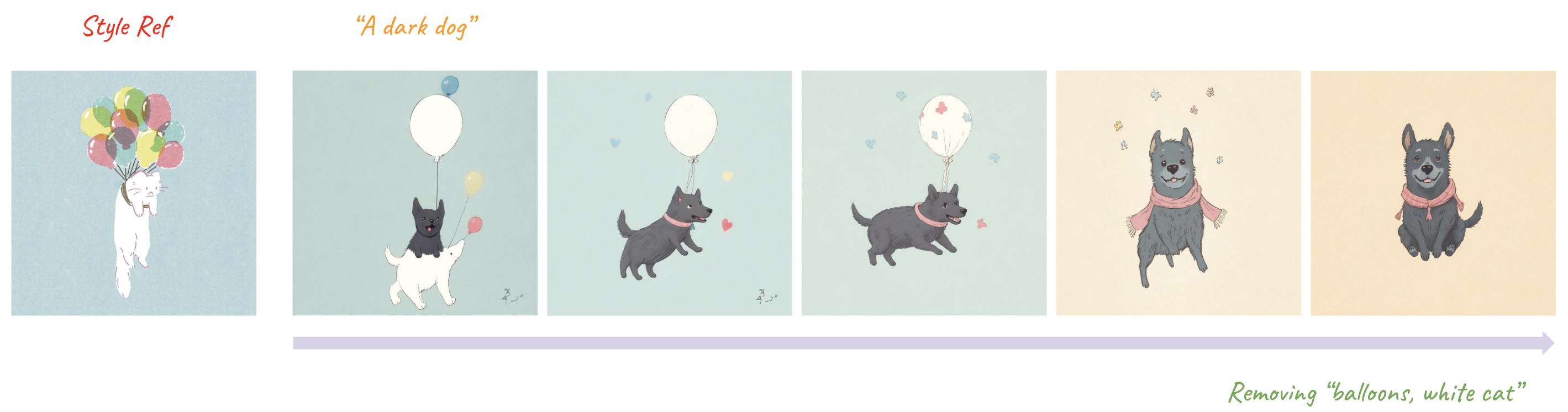}
  \caption{\textbf{The effect of subtraction.} Starting with the original IP-Adapter, we systematically remove content by incrementally increasing the scale of subtraction. As we progress with this approach, the issue of content leakage is effectively mitigated.}
  \label{fig:12}
\end{figure}

\begin{figure}[!h]
  \centering
  \includegraphics[width=\textwidth]{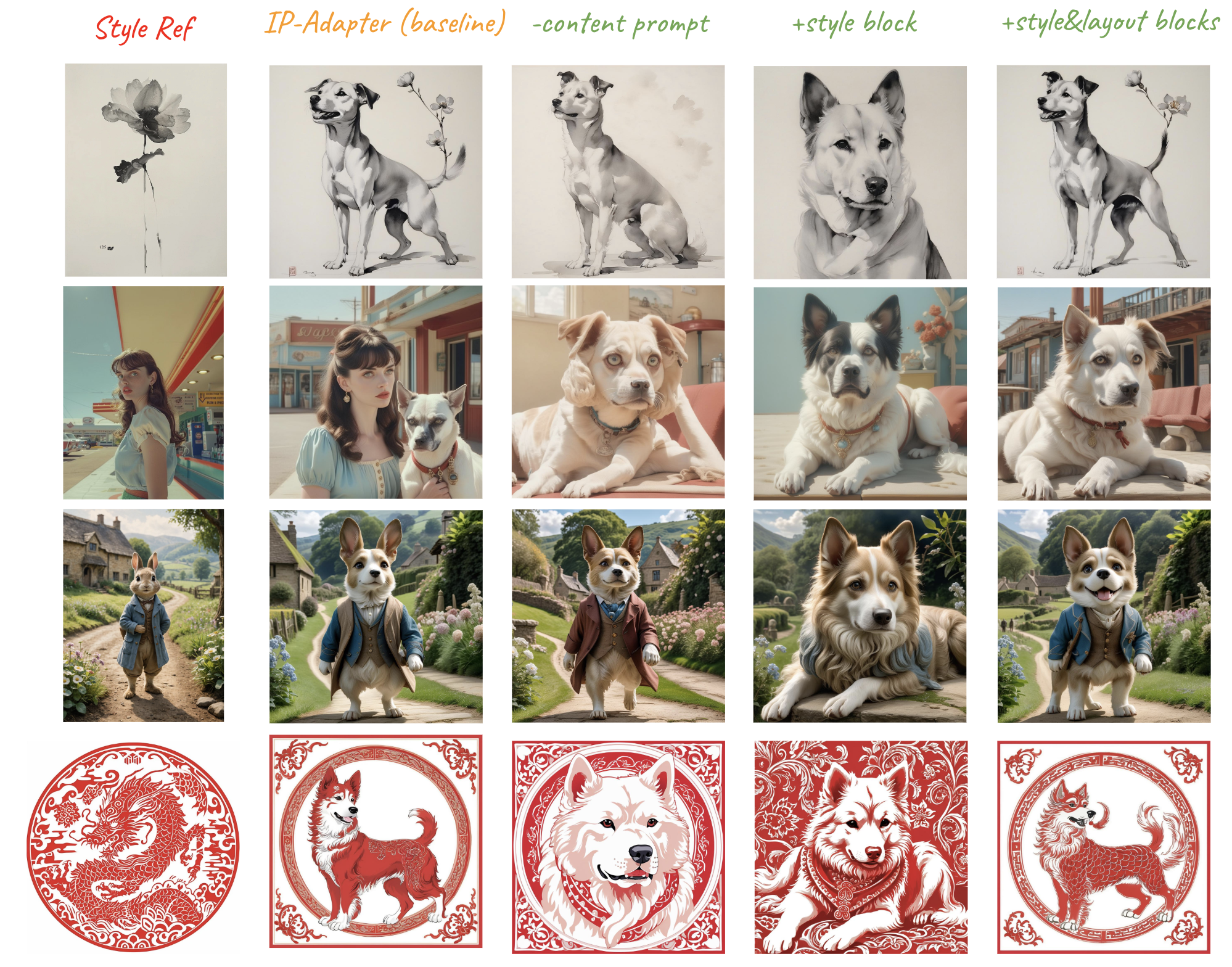}
  \caption{\textbf{The effect of each strategy.} We take the original IP-Adapter as baseline. (1) Subtracting content embedding from image embedding relieves content leakage, but still requires manual weight tuning. (2) Only injecting image features in style blocks performs the best. (3) Injecting image features into both style and layout blocks can handle some special cases where the spatial composition is also a kind of style.}
  \label{fig:9}
\end{figure}

In style transfer, one of the challenges is how to balance style strength and content leakage. For this, we propose two strategies: (1) image embedding subtract content embedding; (2) only inject image features in the style block. In Figure \ref{fig:12}, we show the effect of subtraction. At the beginning without subtraction, although the strength has been manually adjusted, there are still certain content leaks, such as the balloons and white animals in the picture. However, as the intensity of subtraction increases, the consistency between the generated results and the text is significantly enhanced, and there is no longer any content leakage. But it needs to be acknowledged that substraction still requires manual strength tuning.

As shown in the Figure \ref{fig:9}, we found that while maintaining the style as much as possible, reducing the weight of the IP-Adapter still leaves a small amount of content leakage, such as the flowers on the dog's tail in the first row and the female character in the second row, while (1) explicitly removing content elements can avoid content leakage. For (2), we subjectively believe that injecting image features only in the style block is the most concise, elegant, and best-performing way, as shown in the penultimate column. If injected additionally in the layout block, certain content will be leaked, such as the first line. However, it is worth noting that in the third and fourth lines, there may be multiple interpretations of the definition of style, such as whether the anthropomorphic stance in the third line belongs to the style, and whether the round shape of the paper cut in the fourth line belongs to the style. Therefore, we believe that in a specific scene, if the style needs to include spatial layout, it needs to be injected in the layout block at the same time, while for general styles, such as color and atmosphere, only the style block is enough.

%% file: Sections/Conclusion.tex
\section{Conclusions and Future Work}
In this work, we propose InstantStyle as a general framework that
employs two straightforward yet potent techniques for achieving an effective disentanglement of style and content from reference images. Our exploration reveals the unique characteristics of particular attention layers, and our training process from the ground up has demonstrated that not all layers contribute equally, these insights can inspire the training of subsequent models. Both adapter and LoRA approaches can benefit from a reduction in parameter to mitigate overfitting and prevent the leakage of unnecessary information. Moreover, our research findings are directly applicable to other consistent generation tasks that suffer from leakage, and the generation of video generation with consistent stylistic elements. We envision the continuation of our research into consistency generation as a priority for future endeavors, aiming to further refine and expand the capabilities of InstantStyle.